\documentclass[10pt,twocolumn,letterpaper]{article}

\usepackage{cvpr}              

\usepackage{graphicx}
\usepackage{amsmath}
\usepackage{amssymb}
\usepackage{booktabs}

\usepackage[pagebackref,breaklinks,colorlinks]{hyperref}

\usepackage[capitalize]{cleveref}
\crefname{section}{Sec.}{Secs.}
\Crefname{section}{Section}{Sections}
\Crefname{table}{Table}{Tables}
\crefname{table}{Tab.}{Tabs.}

\def\cvprPaperID{007} 
\def\confName{CVPR}
\def\confYear{2022}

\begin{document}

\title{SkateboardAI: The Coolest Video Action Recognition for Skateboarding}

\author{Hanxiao Chen\\
Harbin Institute of Technology\\
Harbin, China\\
{\tt\small hanxiaochen@hit.edu.cn}
}
\maketitle

\begin{abstract}
   Impressed by the coolest skateboarding sport in 2021 Olympic Games, we are the first to curate the original real-world video datasets “SkateboardAI” in the wild, even self-design and implement diverse uni-modal and multi-modal video action recognition methods to recognize different tricks accurately. For uni-modal approaches, we separately apply (1) CNN and LSTM; (2) CNN and BiLSTM; (3) CNN and BiLSTM with the effective attention mechanisms; (4) Transformer-based action recognition pipeline. Transferred to the multi-modal conditions, we investigated the two-stream Inflated-3D architecture on “SkateboardAI” datasets to compare its performance with uni-modal cases. In sum, our objective is developing an excellent AI sport referee for the coolest skateboarding competitions.
\end{abstract}

\section{Introduction}
\label{sec:intro}

Originated in the United States, skateboarding is a popular action sport that involves riding and performing tricks utilizing a skateboard, which can also be considered as an entertainment industry job, a recreational activity, a special art form, and a transportation method. Throughout the years skateboarding has been developed globally, even represented at 2020 Summer Olympics in Tokyo for both street and park-style competitions requiring extraordinary creativity and individuality.

Human action recognition is a standard and well-applied computer vision mission in diverse scenarios containing human-computer interaction and advanced robotics. Furthermore, recognizing diverse human activities from video sequences or still images has been well investigated to develop multiple effective uni-modal and multi-modal approaches. Currently most experimental results are based on normal video datasets including UCF-101, HMDB51, Kinetics, Something-Something V2,  and YouTube-4M. In our work, we focus on action recognition for skateboarding tricks. Whereas, there exists no reasonable datasets for this interesting application, so we are the first to develop the “SkateboardAI” datasets with wild skateboarding videos, then implemented and compared multiple methods to correctly classify video trick data into its underlying category.

Immersed by the rapid growth of tremendous amount of videos, automatic video classification or action recognition is considered as a significant research problem. Most researchers have developed diverse approaches based on advanced deep learning technology. Initially, A. Karpathy et al. \cite{1} proposed multiple solutions to extend the connectivity of CNN in time domain for large-scale video classification, even discovered that a multi-resolution, foveated architecture is much more promising. Different from still images, videos include significant temporal information to be investigated well. For time-series analysis, long short-term memory (LSTM) \cite{2} serves as an effective recurrent neural network (RNN) architecture. Therefore, it’s usually combined sequentially with CNN networks \cite{3} which extract visual features from video frames for action recognition. Z. Wu et al. \cite{4} even employed LSTM on both spatial features and short-term motion features to model longer-term temporal clues for better video classification. Based on this fundamental pipeline, \cite{5} explores more experiments to integrate attention mechanisms into the normal CNN-LSTM benchmarks on UCF-101 and Sports-1M-99 datasets, even achieving better than 0.95 average accuracy with the revised attention block.

Multi-modal approaches which consider multiple types of visual representations (e.g., color-based information, motion-based optical flow) also play an important role in action recognition. To begin with, K. Simonyan and A. Zisserman \cite{6} proposed a two-stream convolutional network architecture including one path operated on video frames and another network on pre-computed optical flow to achieve much promising performance. Following this research tendency, TDD \cite{7}, TSN \cite{8}, etc are proposed to enrich the whole model repository. Consider that 3D convolutions could simultaneously handle both spatial and temporal dimensions, 3D CNN based solutions are deeply investigated and S. Ji et al. \cite{9} designed the special 3D CNN architecture for action recognition. Furthermore, \cite{10} proposed the famous I3D framework which inflates the ImageNet pre-trained 2D model weights to the counterparts in 3D models, even impressively achieved 95.6\% on UCF101 and 74.8\% on HMDB51. Later, R3D \cite{11} and SlowFast \cite{12} appeared following I3D for advantages of 3D convolution kernels.

Inspired by explosive methods in action recognition, we will self-design and implement diverse uni-modal or multi-modal approaches on our curated “SkateboardAI” datasets. Therefore, the remainder of our paper will be organized as follows. {\bf Section 2} presents more details about data collection and analysis for the “SkateboardAI” datasets. Multiple implemented methods for skateboarding recognition will be introduced in {\bf Section 3}. Following this part we present experimental results for each approach and jointly compare their performance in {\bf Section 4}. At last, we provide discussions and future research in {\bf Section 5}.

\section{SkateboardAI Datasets}
\label{sec:formatting}

We developed the original skateboarding video datasets “SkateboardAI” from multiple social media platforms including YouTube, Twitter, Instagram, and BiliBili. Specifically, we choose 15 different fundamental trick classes: \textit{ Ollie, Kickflip, Shuvit, Manual, Hardflip, 50-50 grind, 5-0 grind, Backside 180, BacksideAir, Boardslide, Boneless180, Smithgrind, Benihana, Impossible, Treflip.} Such tricks (Fig. 1) are usually performed in competitions, thus it’s possible to train action recognition models to serve as an excellent AI sports referee which can correctly recognize the skateboarder’s tricks in diverse matches.

\begin{figure}[t]
  \centering
   \includegraphics[width=1.0\linewidth]{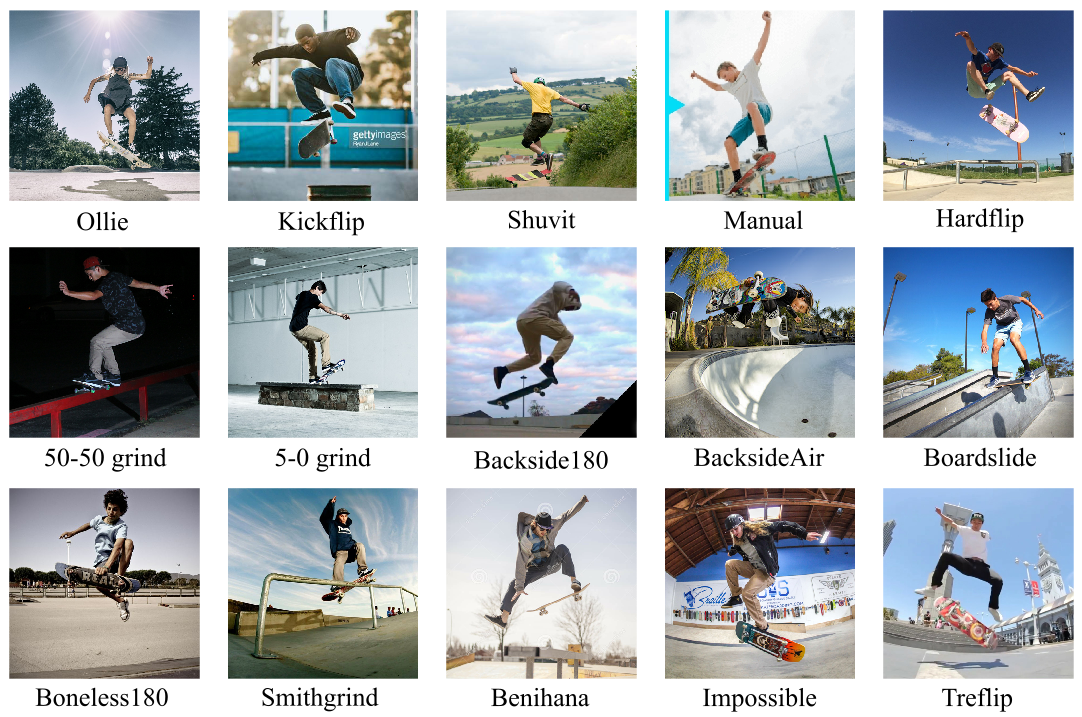}
   \caption{15 different classes in "SkateboardAI".}
   \label{fig:onecol}
\end{figure}

For this action recognition “SkateboardAI” datasets, we collect 50 custom videos for each category and the total number is 750. In this case, we select 45 videos for training and the extra 5 items for validation. Furthermore, we pay attention to two important metrics for datasets: (1) Mean video duration; (2) Video frame number distribution. Fig. 2 represents the mean video duration for whole data, train data, and test data. Actually the maximal trick video time is 10.83s while the minimal is 1.0s in SkateboardAI, so that our collected videos are absolutely lie in the interval of [1, 10] seconds. In addition, the mean duration for “Manual” category videos is the longest whereas the “Kickflip” and “Ollie” are comparatively shorter than others.

\begin{figure}[t]
  \centering
   \includegraphics[width=1.0\linewidth]{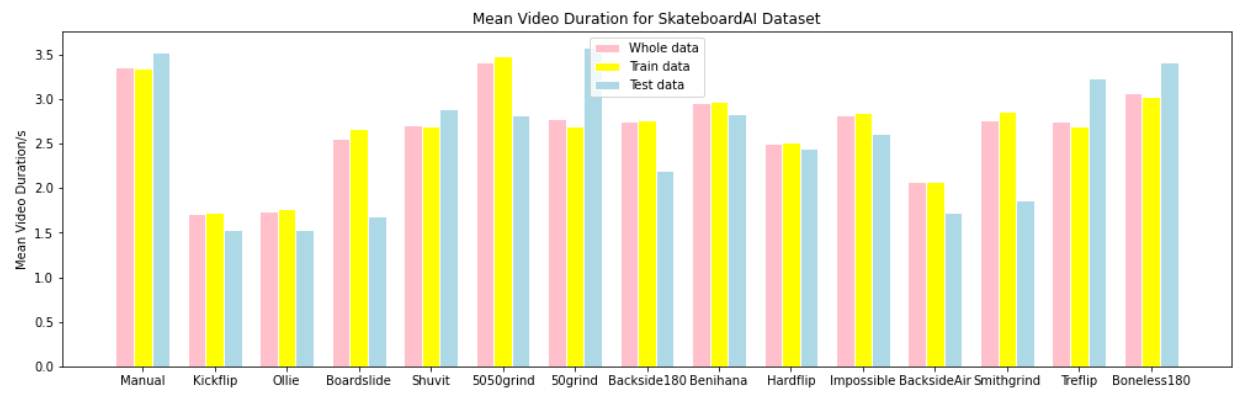}
   \caption{Mean video duration for "SkateboardAI".}
   \label{fig:onecol}
\end{figure}

On the other hand, we even focus on the frame number of whole/train data for model design since we will input certain sequences via cv2.CAP\_PROP\_FRAME\_COUNT. According to Fig. 3, we discover that for whole data 51.07\% videos hold more than 45 and less than 96 frames, 23.47\% videos hold more than 125 frame numbers and the mean frame number for the whole datasets is 96.64. Transferred to train data, the distribution is similar to the whole data that 10.37\% videos hold less than 45 frame numbers and 51.70\% lies in [45, 96].

\begin{figure}[t]
  \centering
   \includegraphics[width=1.0\linewidth]{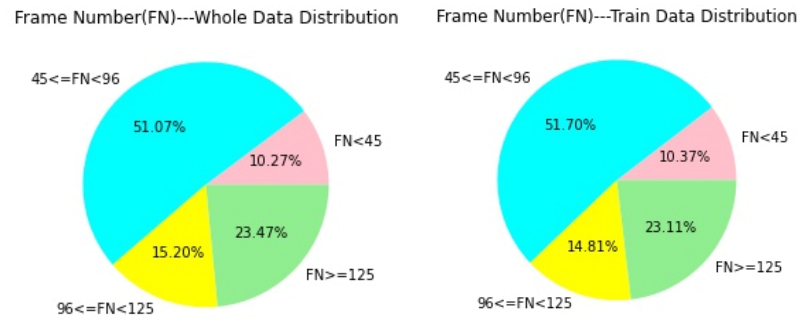}
   \caption{Video frame number distribution for "SkateboardAI".}
   \label{fig:onecol}
\end{figure}

\section{Diverse Approaches}
\label{sec:formatting}
As for addressing action recognition issues on our original “SkateboardAI” datasets, we self-implemented multiple uni-modal and multi-modal approaches. For uni-modal cases, we explored the traditional CNN-LSTM pipeline and its different variants such as CNN-BiLSTM, CNN-BiLSTM-Attention, and CNN-Attention-BiLSTM, even investigated the transformer-based action recognition model. For multi-modal approach, we applied the classical two-stream Inflated 3D for further analysis.

\subsection{CNN-LSTM}

Following the fundamental training pipeline shown in Fig. 4, we firstly use cv2.CAP\_PROP\_FRAME\_COUNT to compute frame number on “SkateboardAI” videos, then apply the reasonable sampling technique to input specific number of video frame sequences (e.g., 45, 60) into the whole architecture. Also, tf.image.resize is applied to uniformly resize the custom video frame size (e.g., (1440, 812)) into certain square size (e.g., (299, 299)) to reduce computation. Therefore, the pre-trained 2D CNN neural networks will extract visual representations via each frame and pass them together into the following LSTM model along the temporal dimension into one index of video category. LSTMs were developed to overcome vanishing gradient and so an LSTM cell could remember context for long sequences to identify spatial-temporal aspect of videos.

\begin{figure}[t]
  \centering
   \includegraphics[width=1.0\linewidth]{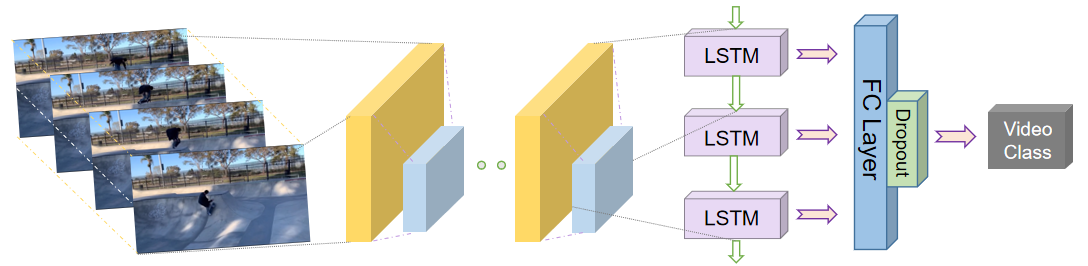}
   \caption{CNN-LSTM action recognition pipeline.}
   \label{fig:onecol}
\end{figure}

\subsection{CNN-BiLSTM}

To comprehensively process input video data, Bidirectional LSTM has been proposed to achieve more promising performance on recognition. As its name means, BiLSTM consists of two LSTMs: one taking the input in the forward direction, and the other in the backwards direction. Considering that adjacent video frames contain strong correlations and the later video frames can be substantially determined or predicted by the previous video shots, BiLSTM is able to preserve all information from both past and future, even effectively increase the amount of useful information available to the neural networks, improving more reasonable context to the algorithm. Thus, Fig. 5 clearly shows our CNN-BiLSTM training pipeline for “SkateboardAI” action recognition which directly changes the LSTM to BiLSTM with the simpler implementation of tf.keras.layers.Bidirectional on LSTM training scripts.

\begin{figure}[t]
  \centering
   \includegraphics[width=1.0\linewidth]{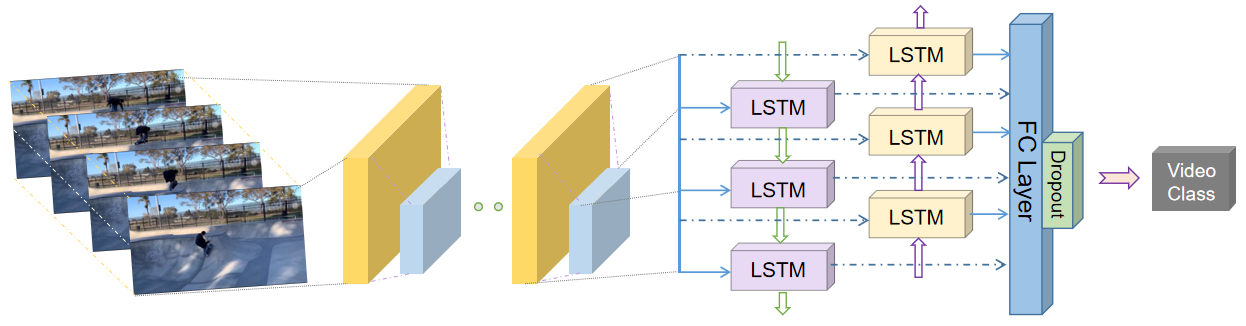}
   \caption{CNN-BiLSTM action recognition pipeline.}
   \label{fig:onecol}
\end{figure}

\subsection{CNN-BiLSTM-Attention}

It’s scientifically investigated that "attention" is a very important mechanism in human visual perception system, which confirms that spatio-temporal features or certain attention segments may profoundly impress and affect our signal perception. Different from the normal CNN-LSTM and CNN-BiLSTM approaches which assume that all temporal representations contribute evenly to the video content classification, we integrate a simple but efficient attention block as \cite{5} into the original CNN-BiLSTM architecture to appropriately apply attention mechanisms for better recognition. In theory, it is possible to add attention on the output of CNNs before LSTM layers, or after LSTM layers that can produce a sequence of outputs. Therefore, a simple attention block can be developed by adding a fully connected layer with an activation function (Softmax) serving as attention probability distribution, which is then multiplied with outputs. Fig. 6 clearly introduces the CNN-BiLSTM-Attention framework that combines the attention block after BiLSTM with two permutation sections. The first permutation is performed on BiLSTM outputs so that the attention block is applied to the dimension of time-steps of video frames rather than image features. Then the output of the attention dense layer is permuted to match LSTM outputs for multiplication. After that, we design to add the fully connected layer followed by a dropout layer to the flattened outputs of the attention block for final category prediction.

\begin{figure}[t]
  \centering
   \includegraphics[width=1.0\linewidth]{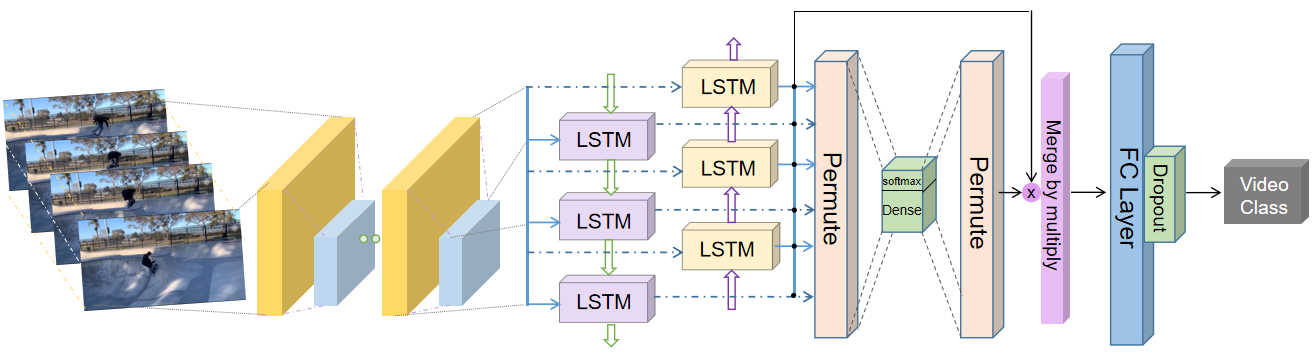}
   \caption{CNN-BiLSTM-Attention action recognition pipeline.}
   \label{fig:onecol}
\end{figure}

\subsection{CNN-Attention-BiLSTM}

Similar to the CNN-BiLSTM-Attention framework, we even investigate to add the attention block between CNN and BiLSTM in Fig. 7. The attention part also consists of two permutation operations and the extracted video frame features are permuted firstly. Subsequently, the attention conditioned information is passed to BiLSTM for data processing and skateboard identification.

\begin{figure}[t]
  \centering
   \includegraphics[width=1.0\linewidth]{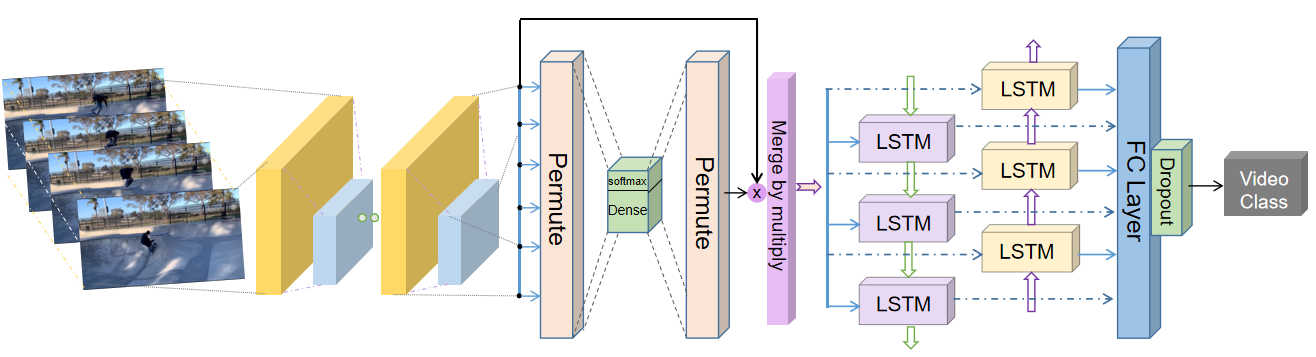}
   \caption{CNN-Attention-BiLSTM action recognition pipeline.}
   \label{fig:onecol}
\end{figure}

\subsection{Transformer-based Method}

Observed that Transformer \cite{13} has gradually become a ubiquitous deep learning approach to achieve outstanding performance on neural machine translation and automatic music generation, we also explore and implement the transformer-based method in Fig. 8 for action recognition on our “SkateboardAI” datasets. 

\begin{figure}[t]
  \centering
   \includegraphics[width=1.0\linewidth]{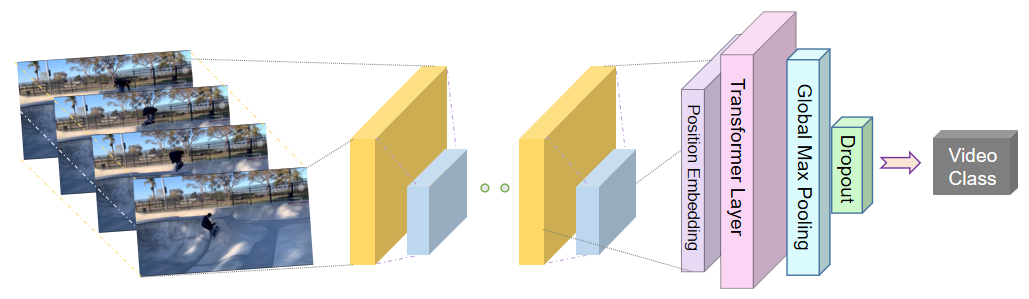}
   \caption{Transformer-based action recognition pipeline.}
   \label{fig:onecol}
\end{figure}

In the first place, we load skateboarding videos by the provided training/validation lists and reduce the custom frame image size (e.g., 1440*806) to 224*224 with the crop\_center or tf.image.resize techniques to speed up computation. Then we apply an ImageNet pre-trained CNN backbone (e.g., DenseNet121, ResNet50) for feature extraction and directly pad each video shorter to certain sequence length. While establishing the transformer-based pipeline, we embed the positions of frames present inside videos with an Embedding layer and add these positional embeddings to the pre-computed CNN video feature maps to make self-attention layers within the Transformer consider video order-agnostic information. After that we employ the Global Max Pooling layer and Dropout to outputs of Transformer model for further video prediction.

\subsection{I3D Multi-modal Method}
Consider that more attention is focused on multi-modal methods in academia and an action event can be described by different types of representations to provide more information, we even investigate the popular I3D \cite{10} multi-modal method to recognize skateboarding tricks. Similar to the official implementation on Kinetics Datasets, we directly feed the whole video into the model rather than the derived video frame RGB images. In fact, the inflated 3D ConvNet (I3D) model is a version of Inception-V1 with batch normalization which has been pre-trained on ImageNet and then "inflated" from 2 dimensions into 3 dimensions. As represented in Fig. 9, I3D utilizes a two-stream architecture with two good modalites of videos: RGB and optical flow. Each stream is separate and the output of models will be combined only at the logit-level for class prediction. To emphasize, the optical flow is computed with the TV-L1 technique aligning with RGB modality for training.
\begin{figure}[t]
  \centering
   \includegraphics[width=0.6\linewidth]{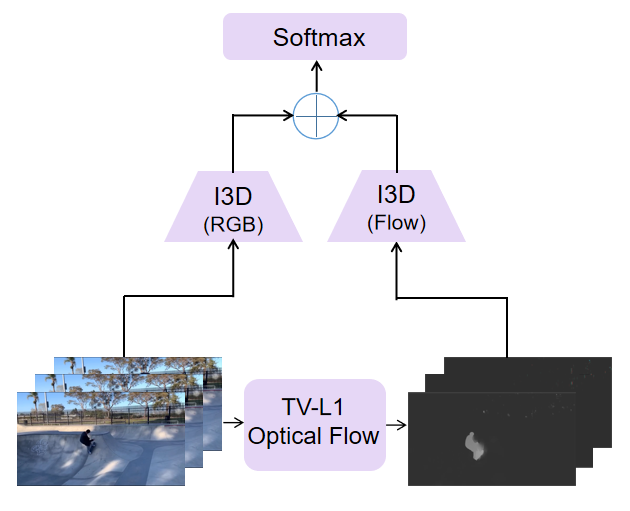}
   \caption{I3D Multi-modal action recognition pipeline.}
   \label{fig:onecol}
\end{figure}

\section{Experiments}
\label{sec:formatting}
As in Section 3, we conducted extensive experiments on diverse models to recognize “SkateboardAI” datasets. Different from previous action recognition methods that firstly extract and save RGB video frame images in the folder to load segments, for each model we directly input and preprocess the whole video stream to reduce training time and efficiently sample frames in the certain sequence length. For CNN-LSTM based approaches consisting of CNN-BiLSTM, CNN-BiLSTM-Attention, and CNN-Attention-BiLSTM, we investigated four different CNN backbones including the ResNet50, ResNet152, DenseNet and VGG16, even set the sequence length as 45 and resize the video image to be (299,299) for 100 training epochs, then we collected three important metrics for each condition: \textit{Training time, Training accuracy, and Validation accuracy.} Table 1 presents the whole numerical results for variants of CNN-LSTM related methods. Obviously, we discover that ResNet50-Attention-BiLSTM achieves the best performance on “SkateboardAI” with 84\% validation accuracy in just 3967.73s training time. Compared with ResNet50-LSTM \& ResNet50-BiLSTM, our integrated attention mechanism can exactly improve the classification accuracy and the attention block can perform much better before BiLSTM instead of after it. Based on comprehensive analysis, ResNet50 and ResNet152 as the pre-trained CNN backbone can extract more effective visual features for LSTM than DenseNet121 or VGG16. As for the training time, VGG16-BiLSTM can just achieve 2347.7s whereas ResNet152 based approaches require more than 6100s due to much more complicated model architectures.

\begin{table}
  \centering
  \scalebox{0.8}{
  \begin{tabular}{@{}lccc@{}}
    \toprule
    Method & Train time/s & Train\_acc & Val\_acc\\
    \midrule
    ResNet50+LSTM \color{blue}{(2048)} & 3676.15 & 0.9956 & 0.8000 \\
    ResNet50+BiLSTM & \bf 3593.21 & 1.0000 &	0.8133 \\
    ResNet50+Attention+BiLSTM &	3967.73 &	1.0000 & \bf 0.8400 \\
    ResNet50+BiLSTM+Attention & 3916.80 &	0.9926 & 0.8133 \\
    ResNet152+LSTM \color{blue}{(2048)} &	6145.99 &	0.9926 & 0.7867 \\
    ResNet152+BiLSTM & 6443.21 & 0.9985 &	0.8000 \\
    ResNet152+Attention+BiLSTM & 6422.44 &	1.0000 & 0.8133 \\
    ResNet152+BiLSTM+Attention & 6372.50 &	0.9985 & 0.7733 \\
    DenseNet121+LSTM \color{blue}{(1024)} & 2881.21 &	0.9822 & 0.7467 \\
    DenseNet121+BiLSTM & 3536.98 & 1.0000 &	0.7067 \\
    DenseNet121+Attention+BiLSTM & 3067.68 & 0.9970 & 0.8000 \\
    DenseNet121+BiLSTM+Attention & 2817.63 & 1.0000 & 0.6933 \\
    Vgg16+LSTM \color{blue}{(512)} & 2405.80 & 0.9867 &	0.7067 \\
    Vgg16+BiLSTM & 2347.70 & 1.0000 & 0.6933 \\
    Vgg16+Attention+BiLSTM & 3247.76 &	1.0000 & 0.6533 \\
    Vgg16+BiLSTM+Attention & 3324.39 &	0.9970 & 0.7867 \\
    \bottomrule
  \end{tabular}}
  \caption{Experimental Results.}
  \label{tab:example}
\end{table}

In addition, based on the Tensorflow platform we also record qualitative results for 16 different training modes: (1) Epoch accuracy and loss visualization results through Tensorboard. (Fig. 10) (2) Dot plotting validation  pictures for 15 classes of test datasets (Fig. 11) to indicate the probability that the trained model recognizes each test video. Furthermore, we saved 16 CNN-LSTM based H5 models and evaluate them on whole “SkateboardAI” datasets or other new collected videos with the predicted category. More experimental and validation results can refer to this link \footnote{https://pan.baidu.com/s/1x0rEGzpIb9HJj7zN5OBJ7w. (key: cscy)}.

\begin{figure}[t]
  \centering
   \includegraphics[width=1.0\linewidth]{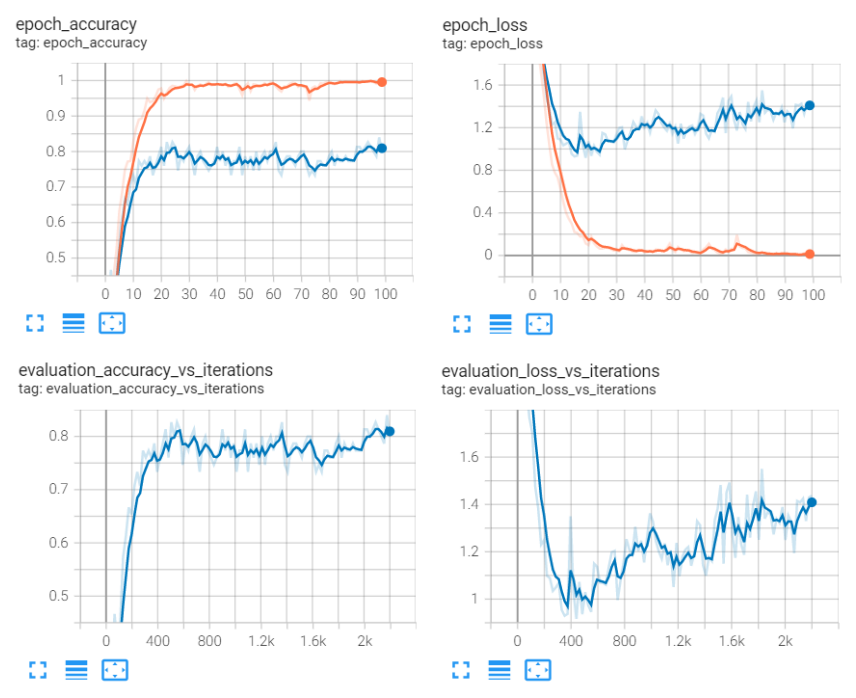}
   \caption{Epoch accuracy and loss visualization for the training ResNet50+LSTM case.}
   \label{fig:onecol}
\end{figure}

\begin{figure}[t]
  \centering
   \includegraphics[width=1.0\linewidth]{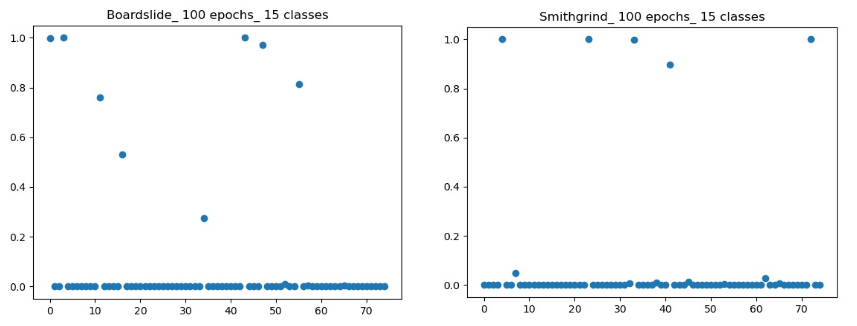}
   \caption{Dot plotting validation sample results (e.g., Boardslide, Smithgrind) for “SkateboardAI” 15 classes after training.}
   \label{fig:onecol}
\end{figure}

Since ResNet50-Attention-BiLSTM is the best pipeline for “SkateboardAI” action recognition among 16 cases in Table 1, we even employ different parameters (e.g., sequence length, video frame size, training epochs) to investigate more ablation study. According to Table 2, we compare three metrics for each case and summarize the following conclusions: (1) 500/200 training epochs may perform worse than 100 epochs training results. (2) Inputting 60 sequences video segments represents a little bit lower performance for the pipeline with 45 sequence length. (3) No matter the smaller image size (224, 224) or the bigger one (366, 366), their validation accuracy is lower than 80\% instead of 84\% for (299, 299). Thus, we discover that more training epochs and more video sequences may not contribute to the original “ResNet50-Attention-BiLSTM” (100 epochs with 45 seq\_length) architecture for much better performance.

\begin{table}
  \centering
  \scalebox{0.76}{
  \begin{tabular}{@{}lccc@{}}
    \toprule
    Method (frame size (e.g., (299,299))) & Train time/s & Train\_acc & Val\_acc\\
    \midrule
    100epochs, seq\_len = 45, (299,299) & 3967.73 & 1.0000 & \bf 0.8400 \\
    100epochs, seq\_len = 60, (299,299) &	5315.06 & 0.9941 & 0.8267 \\
    100epochs, seq\_len = 45, (224,224) &	\bf 2605.73 & 1.0000 & 0.7867 \\
    100epochs, seq\_len = 45, (366,366) &	6222.71 & 1.0000 & 0.7733 \\
    200epochs, seq\_len = 45, (299,299) &	4871.83 & 1.0000 & 0.8133 \\
    500epochs, seq\_len = 45, (299,299) &	7928.08 & 1.0000 & 0.6933 \\
    \bottomrule
  \end{tabular}}
  \caption{Ablation study for "ResNet50-Attention-BiLSTM".}
  \label{tab:example}
\end{table}

Transferred to the transformer-based recognition case, we firstly use pre-trained CNN backbones to extract video features and employ two different image resize techniques including crop\_out or tf.image.resize to reduce computation. Following the pipeline in Fig. 8, we conducted experiments with two CNN backbones including DenseNet121 and ResNet50, even investigate 45 or 100 input sequences for 1000 training epochs. For consistent benchmark, we set the same hyper-parameters in our established models that the dense\_dim equals to 12 and num\_heads are 8. According to Table 3, we discover that all transformer-based approaches require much longer training time with lower validation accuracy than aforementioned CNN-LSTM methods. Surprisingly the best test accuracy is just 24\% and more input video sequences fail to achieve better performance. To begin with, we consider that “crop\_center” technique may abandon or ignore significant visual information to cause lower accuracy, so we apply the “tf.image.resize” to alternatively reserve intact frame information and collect more video sequences for training. However, this operation may not be very effective for transformer-based approaches since the ResNet50 revised case just achieves 8\% test accuracy within 68310.9s. In addition, we saved the training checkpoints for video validation and “ResNet50 + 224 (crop\_center) + seq\_len 45” method model qualitatively evaluates Smithgrind34.mov to be (1) Treflip: 19.78\%, (2) Boardslide: 7.66\%, (3) Smithgrind: 15.99\%. In conclusion, transformer based approaches perform indeed poorly on skateboarding action recognition compared with much simpler CNN-LSTM benchmarks.

\begin{table}
  \centering
  \scalebox{0.68}{
  \begin{tabular}{@{}lcccc@{}}
    \toprule
    Methods (seq\_len: 45 or 100) & Train time/s & Train\_acc & Val\_acc & Test\_acc \\
    \midrule
    DenseNet121 + 224 (crop\_center) + 45 & 17936.21 & 1.0000 & 0.3529 & 6.67\% \\
    ResNet50 + 224 (crop\_center) + 45 & 36845.02 & 1.0000 & 0.2647 & 24.00\% \\
    DenseNet121 + 224 (tf.resize) + 100 & 34133.97 & 1.0000 & 0.3333 & 12.00\% \\
    ResNet50 + 224 (tf.resize) + 100 & 68310.90 & 1.0000 & 0 & 8.00\% \\
    \bottomrule
  \end{tabular}}
  \caption{Transformer-based Method Results.}
\end{table}

To investigate multi-modal related approaches, we conducted experiments with the proposed I3D architecture that combines separate RGB and optical flow streams to employ multi-modal information for better recognition. Fig. 12 introduces the I3D training graph via Tensorboard. However, its experimental results with extremely longer training time to co-process both RGB and optical flow for each video, and the lowest validation accuracy in Table 4 can not meet our anticipations to address “SkateboardAI” action recognition issues well as CNN-LSTM with attention mechanisms based on the similar computation platform.

\begin{figure}[t]
  \centering
   \includegraphics[width=1.0\linewidth]{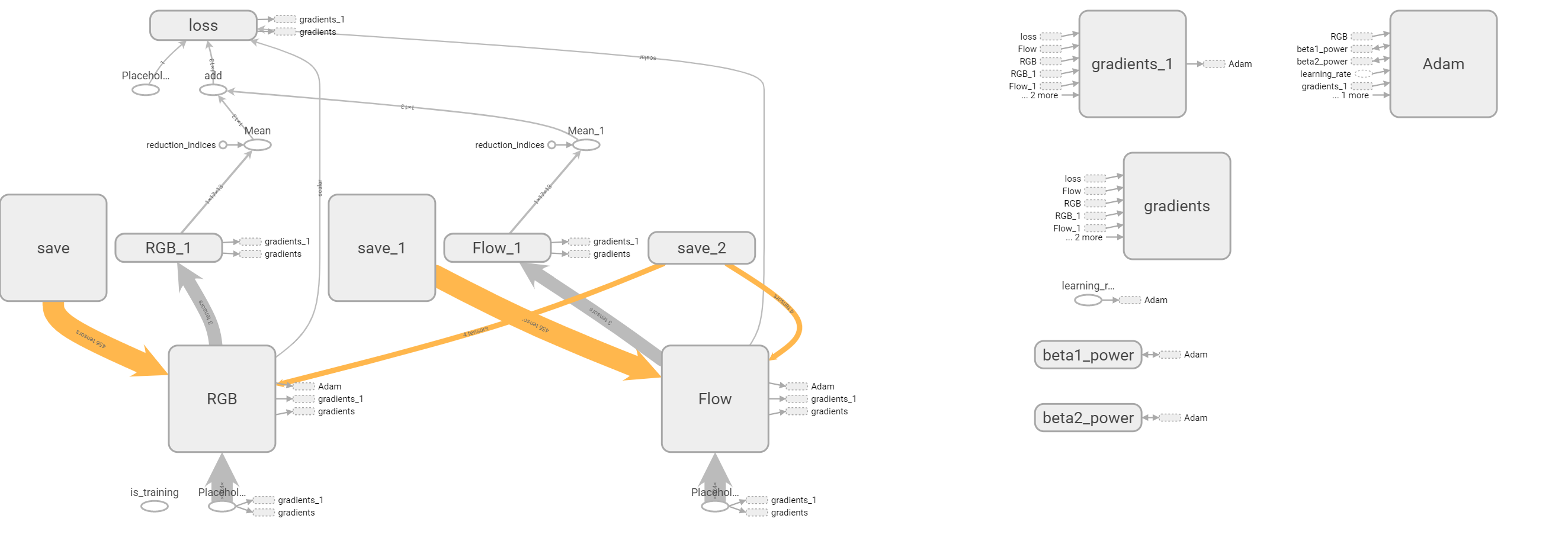}
   \caption{I3D Tensorboard Training Graph.}
   \label{fig:onecol}
\end{figure}

\section{Conclusions}

In summary, we developed the original action recognition datasets “SkateboardAI” with 15 different while fundamental action tricks. And we’ve implemented multiple uni-modal and multi-modal approaches to explore skateboarding identification. To emphasize, our implemented ResNet50+Attention+Bilstm pipeline can achieve the best 84\% validation accuracy than I3D and transformer cases. In future, we’ll expand the datasets and explore it on semi-supervised/unsupervised learning scenarios.

\begin{table}
  \centering
  \scalebox{0.76}{
  \begin{tabular}{@{}lccc@{}}
    \toprule
    Epoch\_num & Train\_acc (mean) & Val\_acc & Train time/s\\
    \midrule
    10 epochs & 0.072 & 0.081 & 432166.05 \\
    50 epochs &	0.074 & 0.078 & 1721886.04 \\
    \bottomrule
  \end{tabular}}
  \caption{I3D Multi-modal Results.}
  \label{tab:example}
\end{table}

{\small
\bibliographystyle{ieee_fullname}
\bibliography{egbib}
}

\end{document}